# Vision-Based Adaptive Robotics for Autonomous Surface Crack Repair

**Joshua Genova[1], Eric Cabrera[2], and Vedhus Hoskere, Ph.D.[1,3]**

[1]Department of Electrical and Computer Engineering, University of Houston; email: jtgenova@cougarnet.uh.edu
[2]Department of Mechanical Engineering, University of Houston; email: eacabre2@cougarnet.uh.edu
[3]Department of Civil and Environmental Engineering, University of Houston; email: vhoskere@central.uh.edu

## ABSTRACT

Surface cracks in infrastructure can lead to severe deterioration and expensive maintenance if not efficiently repaired. Manual repair methods are labor-intensive, time-consuming, and imprecise. While advancements in robotic perception and manipulation have progressed autonomous crack repair, three key challenges remain: accurate localization of cracks in the robot's coordinate frame, adaptability to varying crack sizes, and realistic validation of repairs. We present an adaptive, autonomous robotic system for surface crack detection and repair using advanced sensing technologies to enhance precision and safety for humans. A laser scanner is used to refine crack coordinates for accurate localization. Furthermore, our adaptive crack-filling approach outperforms fixed-speed techniques in efficiency and consistency. We validate our method using 3D-printed cracks under realistic conditions, demonstrating repeatable testing. This research contributes to the field of human-robot interaction by reducing manual labor, improving safety, and streamlining maintenance operations, ultimately paving the way for more sophisticated and integrated construction robotics.

## INTRODUCTION

In the field of infrastructure maintenance, the efficient detection and repair of surface cracks represents one of the most persistent and challenging problems. Although these cracks are often non-structural, they can allow moisture and chemicals to penetrate, leading to gradual deterioration, propagation of damage, and eventually costly repairs or catastrophic failures. Traditional methods for crack repair, such as pouring, filling, sealing, pressure pouring, and banded digging-patching (Gong et al. 2023), rely heavily on manual, resulting in inconsistent quality, safety risks, and prolonged downtime. For example, between 2016 to 2018, surface crack repair in San Francisco International Airport runways cost nearly half a million dollars directly, and millions more in economic impact from the 1,103 flights cancellations and the 13,217 flights delays during repair days (Stock et al. 2019). Similarly, in the aftermath of earthquakes such as the 2011 Christchurch earthquake, inspection and repair efforts can take weeks, months, or even years due to manpower shortages and inaccessibility (Khakurel et al. 2023). These statistics emphasize the noteworthy benefits that could be unlocked by timely, efficient, and proactive surface crack detection and repair that prevent such economic losses and disruptions.

Recent advancements in crack detection and measurement have introduced a variety of innovative systems aimed at enhancing automation and accuracy of crack quantification. (Hoskere 2020) discussed approaches for development of datasets for automated identification damage such as cracks using deep learning. Research has also focused on infrastructure inspection deploying high-resolution sensors and cameras, frequently mounted on automated aerial vehicles (UAVs), automated guided vehicles (AGVs), or robotic manipulators (Khan et al. 2023; Spencer et al. 2019; Yu et al. 2024). Additionally, many researchers have concentrated on computer vision and deep learning approaches for the detection and segmentation of cracks (Hoskere et al. 2020; Xu et al. 2023). Multi-sensor systems, utilizing combinations of RGB-D cameras, laser scanners, and LiDAR, have achieved improved accuracy in quantifying sub-millimeter cracks (Bae et al. 2023). For instance, (G. Alamdari et al. 2024) presented a multi-sensor robotic approach that combines convolutional neural networks, laser scanning, and LiDAR-based 3D point cloud to detect and quantify hairline cracks in concrete structures, capable of measuring crack widths smaller than 0.01 mm. While these studies have demonstrated success in automated inspection, crack detection, and measurement, the logical progression is to move from detection to the actual repair of cracks. In detection, pinpointing the crack center in submillimeter accuracy is not critical, but effective repair relies on accurately mapping cracks within the robotic system's coordinate frame, a challenge that remains largely unaddressed. For example, (Chen et al. 2022) developed a semi-autonomous, RGB-D camera-guided crack filling system that still incurred errors exceeding tens of millimeters, requiring manual intervention. Integrating advanced perception systems for precise crack localization is therefore essential for enhancing the effectiveness and reliability of autonomous repair solutions.

In the realm of robotic crack repair, various advanced tools and end-effectors have been developed to fill cracks of known size and location. (Awuah 2022) modified a 3D printer to be able to dispense bitumen to repair cracks in asphalt concrete in a lab setting. In a similar manner, the Self Repairing Cities project utilized 3D printing technology on a UAV by equipping it with a nozzle mounted on a delta arm, adapted from a commercial 3D printer (Torbaghan et al. 2020). In addition to these approaches, robotic end-effector designs offer promising solutions for autonomous crack repair for more accessible structures (Pereira et al. 2023; Yang et al. 2022). While these methods have made significant advancements, they would not be able to account for cracks of varying widths and depths, resulting in filling errors for variable crack sizes often observed in the real world. (Zhu et al. 2019) designed a manipulator capable of adapting to varying crack depths, although their system was limited to measuring height differences between the end-effector and the crack, without addressing the actual crack-filling process, and does not account for varying crack widths. Implementing sensor feedback to control the robotic system's material flow could enable precise and effective crack repair across diverse geometries.

Evaluating crack filling procedures is challenging because it requires a complete robotic system, consistent filling material, and a repeatable test environment. Consequently, researchers have relied on analogue validations. For example, (Veeraraghavan et al. 2024) validated a mobile robot with a base nozzle using paint droplets on drawn cracks, (Schaefer et al. 2024) tested a syringe-like end-effector on a simulated crack created by taping mortar halves together and manually filling the gap, and (Rahman et al. 2021). developed an AGV-mounted robotic arm whose repair sequence simulated motion without actual material deposition. Although these efforts represent significant progress, realistic validation of crack filling remains insufficient, emphasizing the need for a repeatable test bed that assesses both the repair process and the quality of material deposition.

In summary, despite progress in robotic perception and manipulation, three key technical challenges persist: (i) accurate localization of cracks within the robot's coordinate frame, (ii) adaptability of the filling process to account for variations in crack depth and width, and (iii) validation of the crack repair process through a controlled and repeatable testing environment. Accurate 3D localization is essential for the robot to reach crack centers and execute filling accurately, while adaptability is crucial to prevent inadequate filling, poor adhesion, excess material waste, and early deterioration. Moreover, many systems that perform well in simulation lack real-world validation under dynamic conditions such as changing lighting and filler rheology. Our research addresses these challenges by developing a comprehensive, adaptive, and autonomous system that enhances robotic capabilities to ensure reliable repairs in practical applications.

This paper discusses a novel methodology for crack-repair that addresses both crack detection and repair with high precision. The three novel contributions of our research are:

i. Novel computer vision pipeline that enables accurate crack localization in the robot coordinate system and simultaneous crack measurement using RGB-D camera-laser scanner setup
ii. An autonomous and adaptive crack filling system using a robotic arm and custom end-effector that integrates detection, measurement, and repair for various crack sizes.
iii. A repeatable validation procedure using a 3D printed crack that compares scans before and after filling.

To demonstrate our contributions, we conduct tests in both simulation and in the laboratory environment by constructing the proposed system.

## PROPOSED FRAMEWORK

This section outlines our proposed system and methodology for crack repair, which consists of four parts: extrusion calibration, crack coordinates extraction, crack profile, and crack filling as shown in Fig. 1. First, we calibrate material extrusion by printing at different speeds to map deposition rates. Next, a pre-trained image model segments the crack, extracts its centerline, and converts the pixels into robot coordinates. A laser scanner then refines these coordinates and measures the cross-sectional area to adjust the robot's speed. Finally, material is pumped at a constant flow rate from a custom 3D-printed pump as the robot follows the calculated path, and the repair is validated by rescanning the filled crack and applying heuristic algorithms.

The calibration process ensures accurate material extrusion during crack repair. Despite challenges from varying filler rheology, caused by factors such as reused material, temperature, and humidity, we maintain consistency by adjusting proportions and reusing material to reduce costs. To determine the optimal extrusion rate at different speeds, we print a test strip and use a laser scanner to measure its cross-sectional profile. A heuristic algorithm then detects rapid depth changes to identify the edges, allowing us to calculate the cross-sectional area and central profile of the print. This method guarantees consistent and reliable material deposition despite inherent variations in the mixture.

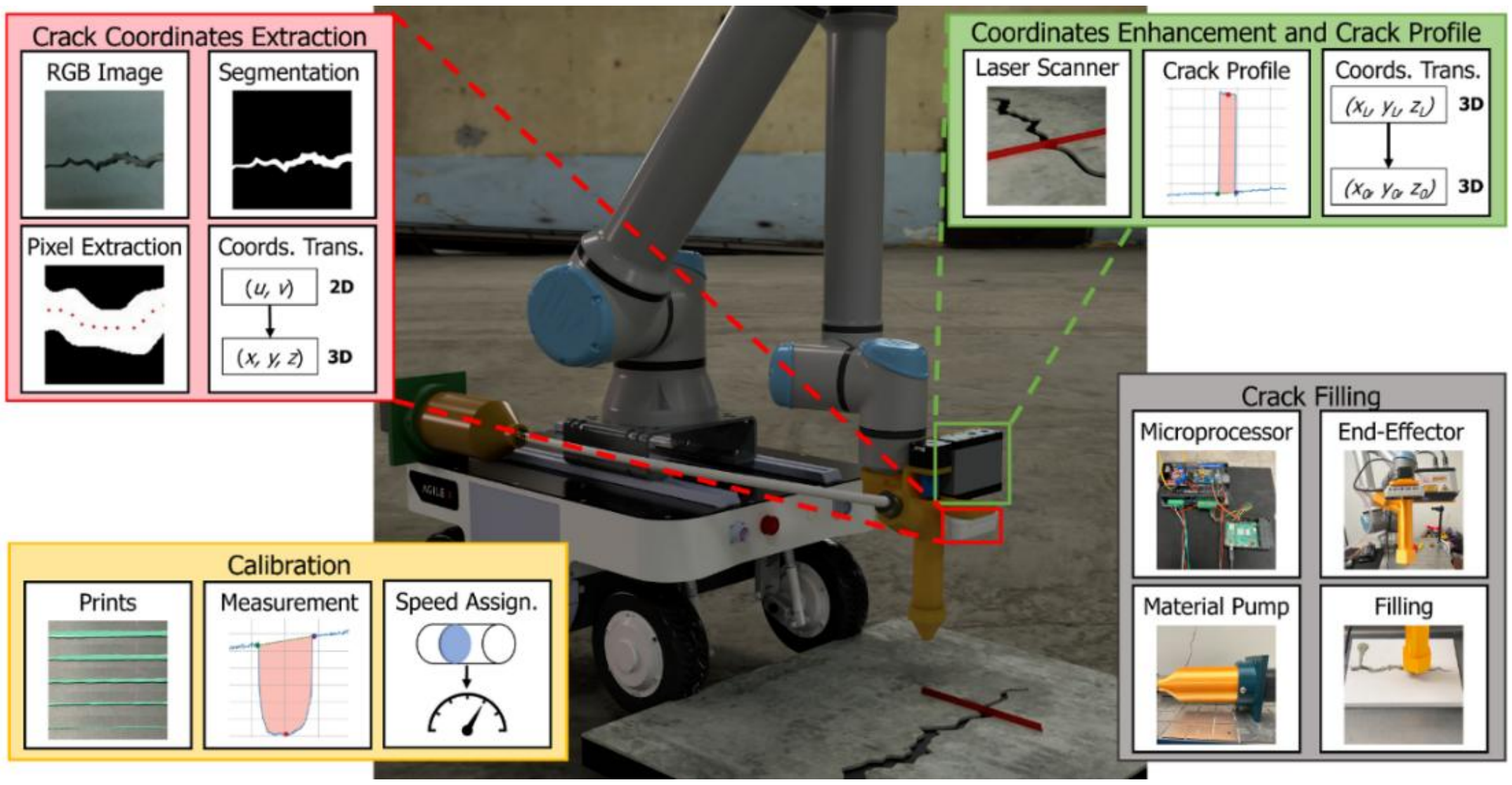

**Fig. 1.** High-Level System Overview.

After calibration, 3D crack coordinates are extracted to allow for automated navigation of the robot along the crack. In our process, an RGB-D camera captures images from which a segmentation algorithm identifies the cracks and produces a binary mask. Our pipeline is agnostic to the specific crack segmentation method utilized, but in this paper, we apply a pretrained DeepLabV3+ model, trained on the conglomerate crack dataset containing over 10,000 images (Bianchi et al. 2022). The mask is then processed to extract a one-pixel-wide centerline via skeletonization, yielding precise pixel coordinates of the crack shown in Fig. 2. These 2D coordinates are converted into 3D camera coordinates using the pin-hole camera model and the camera's intrinsic parameters, given by $\begin{bmatrix} x_c \\ y_c \\ z_c \end{bmatrix} = \boldsymbol{K}^{-1} * z_c * \begin{bmatrix} u \\ v \\ 1 \end{bmatrix}, where\ \boldsymbol{K} = \begin{bmatrix} f_x & 0 & p_x \\ 0 & f_y & p_y \\ 0 & 0 & 1 \end{bmatrix}$. Subsequently, the camera coordinates are transformed into the robot's base frame using the transformation $\begin{bmatrix} x_0 \\ y_0 \\ z_0 \end{bmatrix} = \boldsymbol{T_C^0} * \begin{bmatrix} x_c \\ y_c \\ z_c \end{bmatrix}$, where $\mathbf{T^0}_{\mathrm{C}}$ incorporates the camera's extrinsic parameters. Finally, a heuristic path optimization algorithm minimizes travel distance and avoids redundant movements, resulting in an efficient and reliable navigation path for the crack repair process.

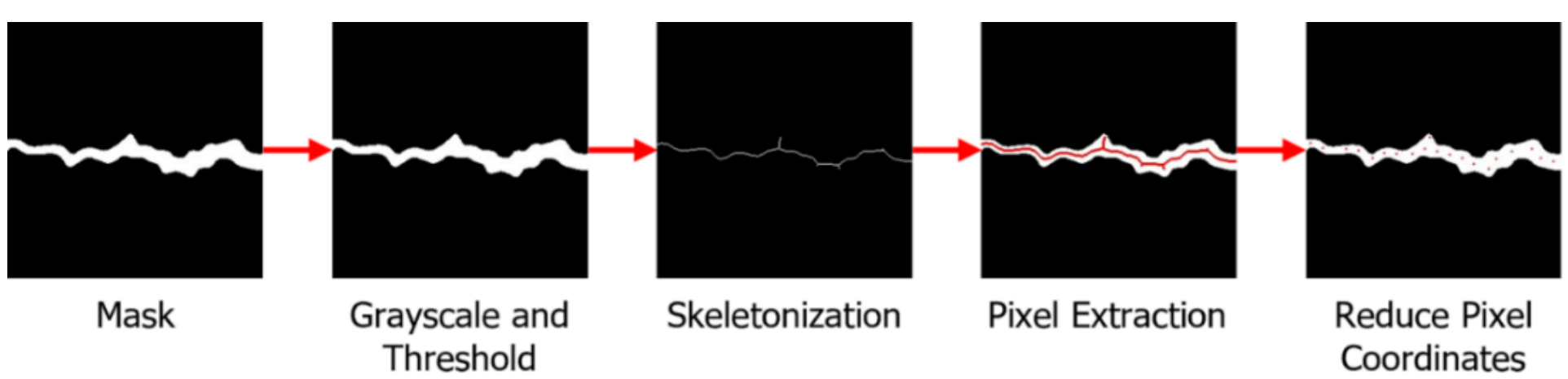

**Fig. 2.** Pixel Coordinate Extraction Process.

We enhance the initial, image-based crack coordinates and generate accurate crack profiles using a laser scanner. Although the RGB-D camera provides valuable initial data, its depth measurements can introduce significant 3D localization errors. In contrast, the laser projects a line onto the surface and captures high-resolution 2D depth profiles along the crack at regular intervals, yielding precise depth information necessary for computing the crack's cross-sectional area and geometry. This enhanced data not only improves coordinate accuracy but also determines the optimal robot speed for material extrusion. Finally, the laser-based coordinates are transformed into the robot's base frame, ensuring precise robotic navigation and material deposition.

With enhanced crack profiles and accurate coordinates from both the sensors, the robot begins crack filling using a custom-designed extruder that pumps repair material along the crack. An initial small extrusion clears the nozzle and establishes a consistent flow before the actual repair begins. As the robot traverses an optimized path, its speed is adjusted based on the local cross-sectional area, slowing down in wider sections to ensure proper filling and speeding up in narrower regions. This controlled deposition results in a crack fill that meets the desired repair specifications, and post-process scans confirm that the final profile closely matches the intended design.

After filling, the laser scanner performs a post-fill scan to measure the repaired crack's cross-sectional profile. Using the same heuristic algorithm as before, we compare the pre-fill and post-fill profiles, where under-filled areas retain depth, and over-filled areas exhibit extra height. Repair accuracy is quantified by the absolute normalized difference $\epsilon_{fill} = \left|\frac{A_{post}}{A_{pre}}\right|$, where $A_{pre}$ represents the cross-sectional area of the crack before filling, while $A_{post}$ represents the cross-sectional area of the filled crack at the exact same coordinate. Lower values of $\epsilon_{fill}$ indicate a repair that closely matches the desired geometry, ensuring quality and reliability in the process.

## SYSTEM DESIGN AND EXPERIMENTAL SETUP

To evaluate the performance and validate the functionality of our automated crack repair system, we utilized both simulation and experimental setups. Simulation was conducted in RoboDK, for its robust Python API, cost-effectiveness, and seamless integration with real robots (RoboDK Inc. 2024). This virtual environment allowed us to test and refine the robot's motion and crack detection processes before real-world deployment. The experimental setup included the Universal Robot's UR10e robotic arm, a custom-designed end-effector equipped with an RGB-D camera and laser scanner, a custom-designed pump system, and 3D-printed crack specimens. By combining simulation with hands-on experiments, we ensured a comprehensive evaluation of the system's capabilities and performance.

The system design integrates key components essential for automated crack repair. The UR10e robotic arm navigates along the crack, guided by accurate 3D coordinates and optimized paths. Fig. 3 illustrates the complete setup and the end-effector design. The custom-designed end-effector contains an Intel RealSense D435 RGB-D camera for initial crack detection, a Micro-Epsilon scanControl 3002 laser scanner for coordinate enhancement and detailed crack profiling, and a print head driven by a Nema 17 stepper motor to precisely extrude the filling material. The pump system, inspired by a medical syringe, utilizes a Nema 34 stepper motor to drive a threaded rod and piston assembly within 3D-printed components. A Raspberry Pi with a motor driver controls the pump's extension and retraction at a constant speed, maintaining consistent material flow. Additionally, 3D-printed crack specimens replicate various real-world crack patterns, allowing us to rigorously test and validate the repair process. Play-Doh mixed with water is used as the repair material for its low cost, reusability, and ease of cleanup during experiments. For uniform mixture consistency, the Play-Doh and water are mixed in a baker's mixer for 30-60 minutes to achieve similar viscosity to field-grade crack sealants before being loaded into the pump. Furthermore, system performance was evaluated in real time. The laser scanner acquires and refines crack coordinates within milliseconds, ensuring adaptive material deposition even under dynamic conditions.

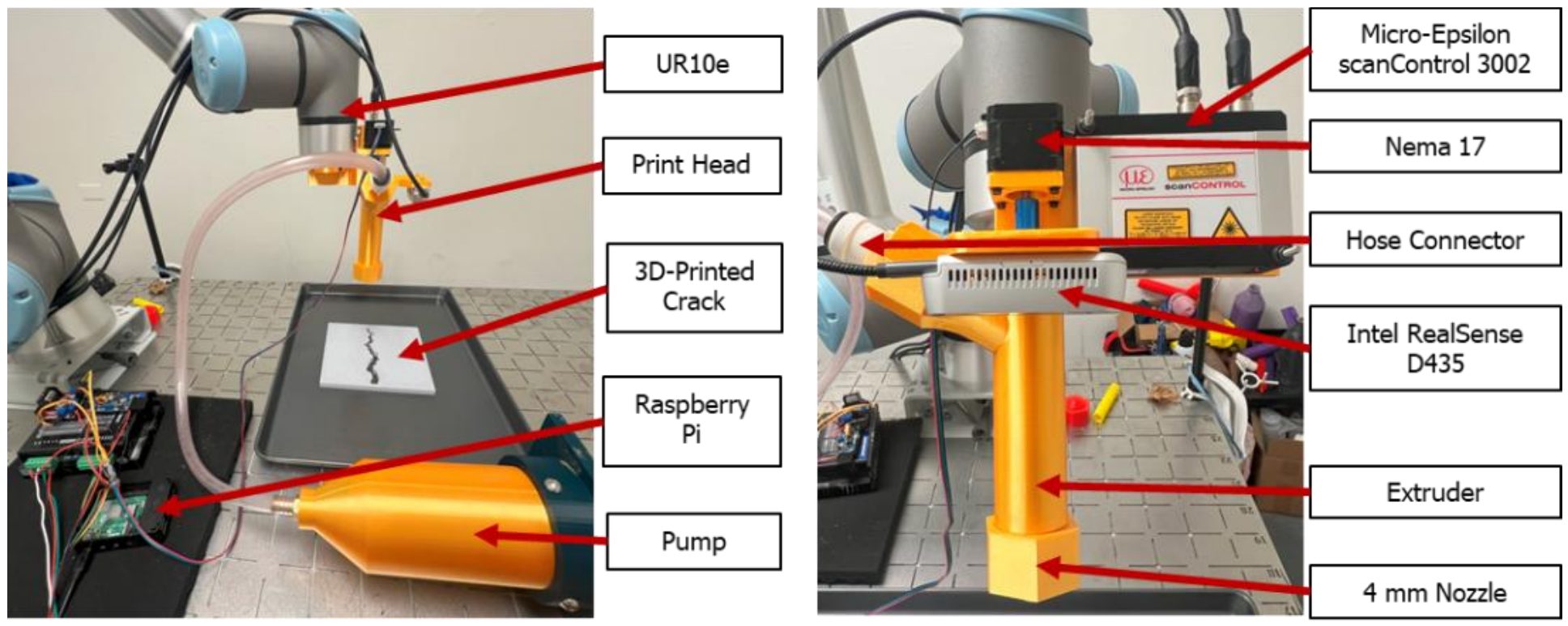

**Fig. 3.** System Design and End-Effector Design

## EXPERIMENTS AND RESULTS

This section presents the results of our automated crack repair system, focusing on material extrusion calibration, crack filling efficiency, and the RGB-D camera vs. RGB-D camera-laser localization accuracy. The calibration results demonstrate the relationship between the robot's speed and material extrusion rate. The validation results assess the accuracy of the crack filling process, comparing pre-fill and post-fill profiles to evaluate the system's effectiveness. The crack localization accuracy proves the laser scanner's ability to improve the crack coordinates, instead of relying solely on the RGB-D camera. In addition, a video link demonstrating the whole crack filling process can be found here.

**Material Extrusion Calibration:** Accurate material extrusion during crack repair relies heavily on a precise calibration process. We printed a 150 mm material strip at a constant robot speed and flow rate, then used a laser scanner to measure the inner 100 mm, avoiding the inconsistent start and end, at 10 mm increments. This approach, based on our heuristic method, allowed us to compute the average cross-sectional area at each interval. By repeating the process at various speeds, we established an inverse relationship: as the robot's speed increases, the extruded material's cross-sectional area decreases, and vice versa as illustrated in Fig. 4. The results, summarized in Table 1 describe how different speeds correspond to distinct cross-sectional areas. This calibration is essential for matching specific speeds to varying crack dimensions, ensuring that each crack coordinate receives the optimal amount of material during the repair process.

**Crack Filling Efficiency**: The efficiency of the crack filling process was evaluated by comparing the pre-fill and post-fill profiles, with error values, $\epsilon_{fill}$ , computed at various speeds, including adaptive speed control as shown in Table 2. The results indicate that adaptive speed control achieved the lowest error or highest accuracy, with a mean $\epsilon_{fill}$ of 0.305, a standard deviation of 0.240, and a median of 0.265. In comparison, fixed speeds of 6, 8, 10, 15, and 20 mm/s resulted in higher mean and standard deviation values, indicating less precise filling with notable overfill or underfill. Fig. 5 demonstrates the qualitative performance of different robot speeds during crack filling, where the adaptive speed shows the least amount of overfill and underfill. Fig. 6 also shows the graph overlay of the crack profile at pre-fill vs. post-fill, which shows the adaptive speed with minimal underfill and overfill. At speed 6 mm/s, there is a large amount of overfill and at speed 20 mm/s, there is a large amount of underfill which is apparent in both Fig. 5 and Fig. 6. The repair process time also varied with speed, with adaptive speed control completing the process in 24.784 seconds, compared to 39.844 seconds for the 6 mm/s speed and 19.102 seconds for the 20 mm/s speed. Although higher speeds reduce process time, they compromise filling accuracy. These findings confirm that the automated crack repair system, applying adaptive speed control, can effectively adjust the material flow based on robot speed and deposition based on crack profile data, ensuring accurate and consistent repairs.

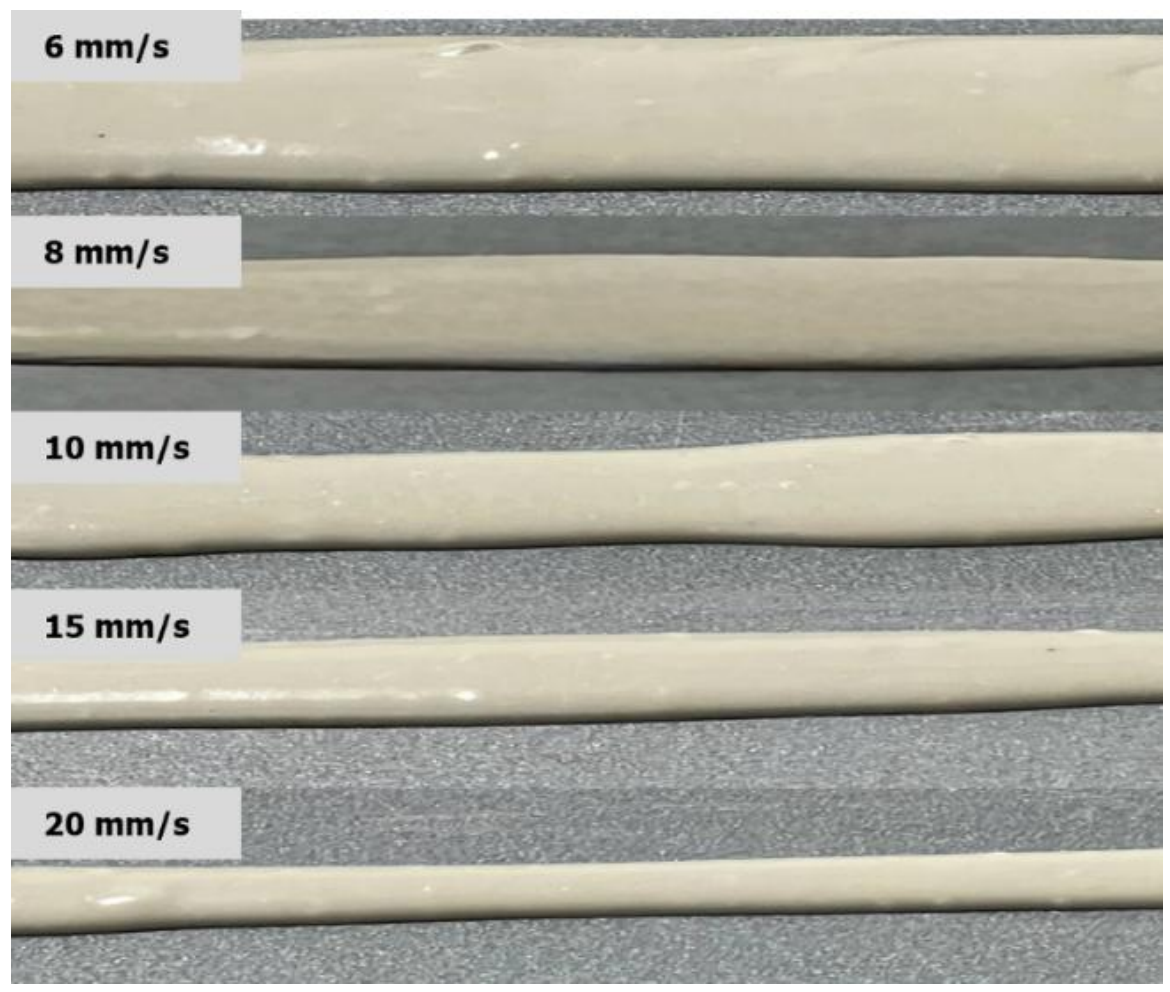

**Fig. 4.** Material Extrusion at Different Robot Speeds.

**Table 1: Robot Speed vs. Cross Sectional Area**

| Robot Speed (mm/s) | Cross-Sectional Area (mm) | Std. Dev. (mm) |
|---|---|---|
| 6 | 165.764 | 18.44 |
| 8 | 111.977 | 10.807 |
| 10 | 91.448 | 8.111 |
| 15 | 63.561 | 7.003 |
| 20 | 41.713 | 7.936 |

**Table 2: $\epsilon_{fill}$ vs. Speed**

| Speed (mm/s) | Mean[1] | Std. Dev. | Median | Time (s) |
|---|---|---|---|---|
| 6 | 2.499 | 3.682 | 1.186 | 39.844 |
| 8 | 1.63 | 2.753 | 0.387 | 31.333 |
| 10 | 1.175 | 2.497 | 0.221 | 28.000 |
| 15 | 0.448 | 0.555 | 0.265 | 21.944 |
| 20 | 0.711 | 0.568 | 0.608 | 19.102 |
| **Adaptive** | **0.305** | **0.240** | **0.265** | **24.784** |

[1]Mean refers to the average from 32 crack coordinates of their $\epsilon_{fill}$

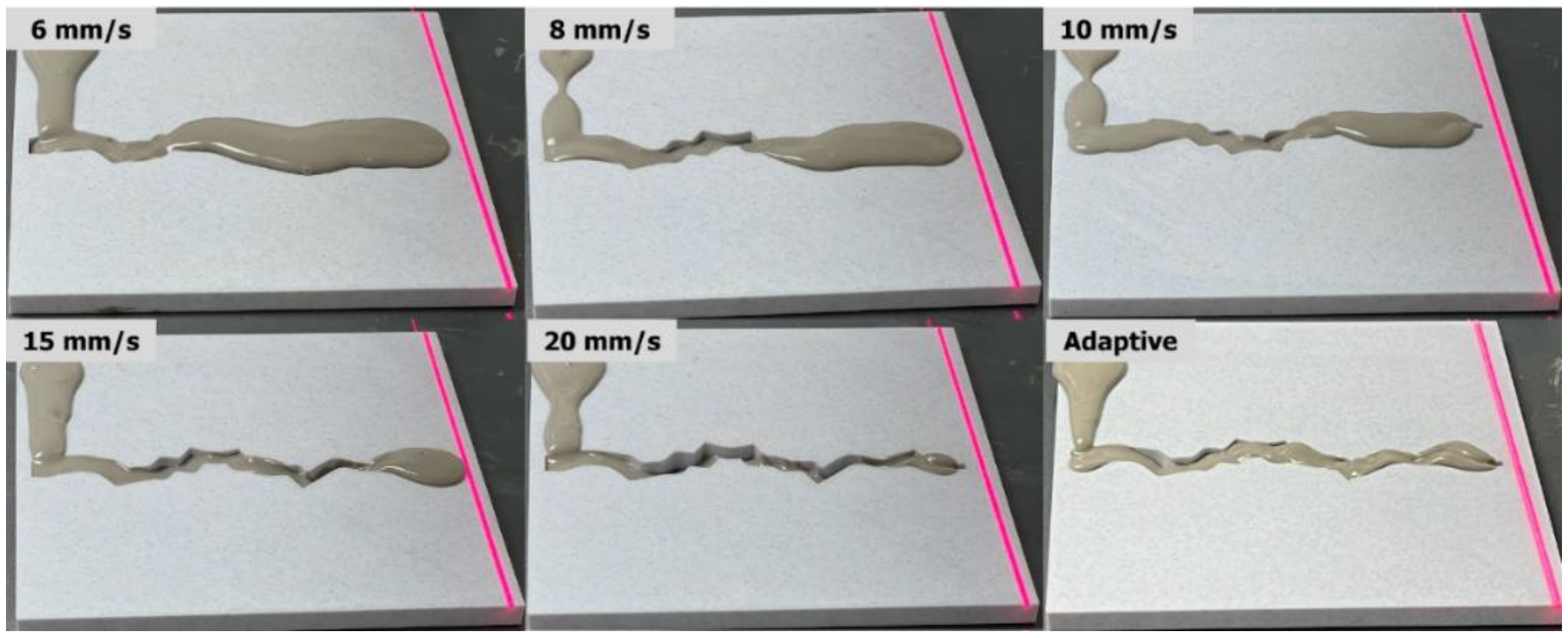

**Fig. 5.** Filled Crack with Different Speeds.

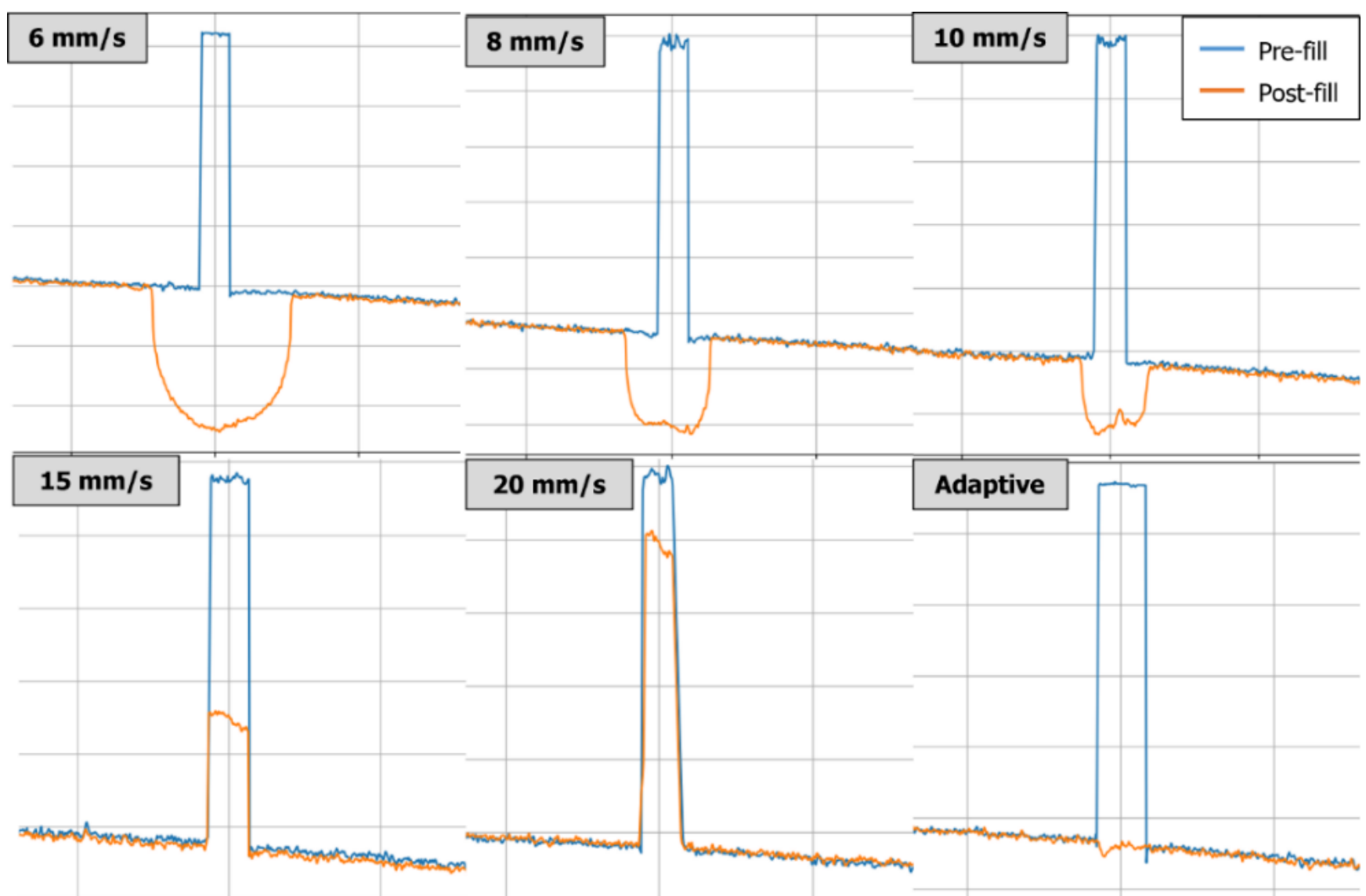

**Fig. 6.** Visual Overlay of Pre-Fill vs. Post-Fill.

**RGB-D vs. RGB-D-Laser Scanner Localization:** To assess the impact of the addition of the laser scanner on the accuracy of crack coordinate localization, we compare the difference in localization with and without the laser scanner. Table 3 describes the average differences in the X, Y, and Z coordinates, along with the overall distance difference, based on 10 complete scans of horizontal cracks, totaling over 240 crack coordinates. The results reveal notable improvements in localization, particularly in the X and Z directions. For instance, errors in the X coordinate were initially large due to the horizontal orientation of the cracks, which could cause the robot to miss the crack entirely, while inaccuracies in the Z coordinate can lead to improper material deposition. In contrast, incorporating the laser scanner significantly reduced these errors, confirming its critical role in ensuring precise crack localization and effective repair.

**Table 3: RGB-D vs. RGB-D-Laser Scanner Crack Coordinates Difference**

| Coordinate | Average Difference (mm) | Std. Dev.(mm) |
|---|---|---|
| X | 13.575 | 2.763 |
| Y | 0.052 | 0.033 |
| Z | 8.364 | 2.432 |
| Distance | 16.134 | 2.740 |

## CONCLUSION

This paper presents a novel design for an adaptive, autonomous system for surface crack detection and repair using a robotic arm equipped with advanced sensing technologies. The system integrates an RGB-D camera for initial crack detection, a laser scanner for precise crack localization in 3D space and measurements, and an extruder for material deposition. Our experiments demonstrate the benefits of our adaptive, autonomous system over fixed-speed methods. Material extrusion calibration established an inverse relationship between robot speed and extrusion rate, enabling adjustments tailored to crack size. Crack filling tests demonstrated that adaptive speed control reduces fill error while optimizing repair time and precision compared to fixed speeds. A comparative study showed that relying solely on an RGB-D camera introduces localization errors of tens of millimeters, whereas incorporating a laser scanner markedly improves accuracy. Finally, using 3D-printed crack specimens and a specially formulated filling material provided a robust test environment, and post-fill validation via laser scanning confirmed the system's effectiveness in achieving accurate repairs. This research shows that adaptive, autonomous systems can greatly improve infrastructure repair by increasing quality and efficiency, enhancing safety, lowering maintenance costs, and extending the lifespan of critical structures. By automating hazardous, repetitive tasks, these systems boost worker safety and allow personnel to focus on complex, judgment-based activities, meeting the growing demand for automated repair solutions.